\documentclass[runningheads]{llncs}
\usepackage[T1]{fontenc}
\usepackage{amsmath,amsfonts}
\usepackage{algorithmic}
\usepackage{graphicx}
\usepackage{textcomp}
\usepackage{xcolor}
\usepackage{xspace}
\usepackage{subfig}
\usepackage{tabu}
\usepackage{tabularx}
\usepackage{multirow}
\usepackage{breakurl}
\usepackage{algorithm}
\usepackage{algorithmic}
\usepackage{xspace}
\usepackage{hyperref}
\usepackage{wrapfig}
\newcommand{\alg}{{Tabula}\xspace}
\newcommand{\algr}{{Tabula\_R}\xspace}
\newcommand{\algp}{{Tabula\_P}\xspace}
\newcommand{\algf}{{Tabula\_F}\xspace}
\newcommand{\algll}{$\text{\alg}_{L}$\xspace}
\newcommand{\algrr}{$\text{\alg}_{R}$\xspace}
\newcommand{\algmm}{$\text{\alg}_{M}$\xspace}
\newcommand{\ctab}{{CTAB-GAN}\xspace}

\newcommand{\ctgan}{{CT-GAN}\xspace}

\newcommand{\ctabplus}{{CTAB-GAN+}\xspace}

\newcommand{\rtf}{{RTF}\xspace}
\newcommand{\great}{{GReaT}\xspace}

\newcommand{\tabddpm}{{TabDDPM}\xspace}

\begin{document}
\title{TabuLa: Harnessing Language Models for Tabular Data Synthesis}


%
%
\author{Zilong Zhao\inst{1,2} \and
Robert Birke\inst{3} \and
Lydia Y. Chen\inst{4}}
\authorrunning{Z.Zhao et al.}
%
\institute{National University of Singapore, Singapore \and Betterdata.ai \\\email{z.zhao@nus.edu.sg} \and
University of Turin, Italy \\\email{robert.birke@unito.it} \and
Delft University of Technology, Netherlands\\
\email{lydiaychen@ieee.org}}

\maketitle              
\begin{abstract}
Tabular data synthesis is crucial for addressing privacy and security concerns in industries reliant on tabular data. While recent advancements adopt large language models (LLMs) for realistic tabular data generation, their long training times and limited reusability hinder practical applications.
In this paper, we propose \alg, a tabular data synthesizer that leverages the structure of LLM. Unlike state-of-the-art (SOTA) LLM-based tabular data synthesizers that rely on pre-trained LLMs, \alg discards the pre-trained weights originally designed for natural language tasks, focusing instead on a tailored approach for tabular data. In addition, \alg introduces a token sequence compression strategy that significantly reduces training time while maintaining data quality, alongside a novel token padding method that improves sequence alignment across training batches. 
Experiments on six datasets show that \alg achieves superior synthetic data utility compared to current SOTA methods. 
Additionally, the results demonstrate that \alg model trained on tabular datasets serves effectively as a foundational model for synthesizing new tabular datasets. 
Furthermore, the proposed padding method outperforms the conventional left and right padding strategies.
Finally, the results highlight that \alg averagely reduces training time per epoch by 46.2\% compared to state-of-the-art LLM approaches while achieving higher data utility. Our code is available at \url{https://github.com/zhao-zilong/Tabula}.

\keywords{LLM  \and Generative Model \and Tabular Data.}
\end{abstract}

\section{Introduction}
While data sharing is crucial for knowledge development, privacy concerns and strict regulations (e.g., European General Data Protection Regulation (GDPR)) limit its full effectiveness.
Since tabular data is a predominant data format, tabular data synthesis has emerged as a critical research area, aiming to generate realistic data while preserving privacy and confidentiality. Prior art has explored this topic using generative adversarial networks (GANs)~\cite{ctgan,ctabgan,ctabplus,itgan,zhu2022}, variational autoencoders (VAEs) ~\cite{ctgan} and diffusion models~\cite{kotelnikov2023tabddpm,sos}. The recent state-of-the-art (SOTA) methods in this domain have leveraged large language models (LLMs)~\cite{borisov2022language,solatorio2023realtabformer} to tackle the challenge of synthesizing tabular data effectively. 

LLMs offers two main advantages comparing to prior SOTAs in tabular data synthesis: (1) The tokenization process of LLMs is fully text-based, eliminating the need to pre-define column data types such as categorical or continuous, which is a requirement for almost all GAN and diffusion model-based tabular data synthesizers; (2) The fully text-based tokenization approach also addresses the dimension explosion problem encountered when using one-hot encoding for high-dimensional data. 
However, this technique faces limitations in training efficiency and preserving cross-column correlations. For instance, the \great~\cite{borisov2022language}, a LLM-based framework, suffers from slow convergence, requiring over 9 hours to achieve synthetic data quality comparable to CTGAN~\cite{ctgan}, which takes just 1 minute~\cite{borisov2022language}.

In response to these challenges, we introduce a novel approach -- \alg, a tabular data synthesizer based on the large language model framework. The goal of \alg is to accelerate the convergence speed of LLM-based methods for tabular data synthesis tasks. We achieve this through four key features:
\color{black}
(i) \textbf{Re-evaluation of pre-trained NLP models for data synthesis}. 
We challenge the traditional reliance on pre-trained natural language processing (NLP) models for tabular data synthesis, as employed by the current SOTA method~\cite{borisov2022language}. Instead, we propose using a randomly initialized or tabular-specific pre-trained model, enabling faster adaptation to tabular synthesis tasks.
(ii) \textbf{Tailoring foundation models for tabular synthesis}.
Our approach focuses on building foundation models optimized for tabular data from scratch, bypassing conventional pre-trained models. This strategy allows for a more efficient learning process, leveraging the unique demands of tabular data synthesis.
(iii) \textbf{Token sequence compression}. 
To better enable the LLM to learn the interdependencies between different tokens and shorten training time, we compress token sequences by condensing column names and categorical values into single tokens. Additionally, we simplify table-to-text representations (e.g., "X is Y" proposed by \cite{borisov2022language} to "X Y"), reducing training time and enhancing the model's ability to learn critical relationships.
(iv) \textbf{Customized token padding strategy}.
In order to achieve consistent token sequence lengths within a training batch, we introduce "Middle Padding", a novel token padding approach for tokenizing tabular data. By placing padding tokens within sequences rather than at the beginnings or ends, this approach guarantees that features within the same data column in the original data maintain identical absolute positions in the newly encoded token sequence, improving tabular data representation and synthesis quality.
\color{black}

Our algorithm undergoes extensive evaluation on six commonly used machine learning (ML) datasets, comprising both classification and regression tasks. 
The results show that a randomly initialized LLM outperforms the conventional pre-trained LLM for tabular data synthesis. 
Additionally, a fine-tuned \alg model on tabular data serves as a strong foundation for new synthesis tasks.
Our token sequence compression reduces training time by 46.2\% compared to SOTA LLM-based synthesizers while still improving synthesis quality.
Furthermore, the middle padding method outperforms traditional left and right padding. 
The main contributions of this study can be summarized as follows:
(1)
Highlight the counter-intuitive result that randomly initialized language models converge faster than well-trained ones for tabular data synthesis. We attribute this to the different tasks: tabular synthesis versus NLP. 
(2) Design an efficient fine-tuning strategy to re-use the previously trained models as a rolling new foundation for new synthesis tasks to improve synthesis quality.
{
(3) Compress token sequence by representing column name and categorical value using single tokens to reduce model training overhead.
}
(4)
Propose \textit{middle padding} strategy designed for the  tabular data representation in LLM.  

\section{Related Work}

There have been various approaches for synthesizing tabular data. Probabilistic models like Copulas~\cite{copulas} use Copula functions to model multivariate distributions but struggle with categorical data. Synthpop~\cite{nowok2016synthpop} generates data variable by variable using regression models, but this method is computationally intensive. Bayesian networks~\cite{privb,avino2018generating} are effective for categorical data but lack support for continuous variables.

Recently, deep generative models such as GANs, VAEs and diffusion models have attracted the attention for tabular data synthesis.
\ctgan~\cite{ctgan}, \ctab~\cite{ctabgan} and \ctabplus~\cite{ctabplus} improve data synthesis by introducing several preprocessing steps for various data types and distributions to encode data into a suitable form for GAN and VAE training. The conditional vector designed by \ctgan and later improved by \ctabplus reduces mode-collapse on imbalanced continuous columns. 
TabDDPM~\cite{kotelnikov2023tabddpm} and SOS~\cite{sos} use diffusion models for tabular data synthesis. TabDDPM separately synthesizes categorical and continuous data, which does not maintain well correlations between categorical and continuous  columns. SOS is specifically designed for oversampling minority class of tabular data. None of the above algorithms allows to generate data conditioned on both categorical and continuous values. In addition, since one-hot encoding is used for categorical data for all above methods. 
it is difficult to synthesize tabular data with high-dimensional categorical columns such as "Zip Code". 
\great~\cite{borisov2022language} and REaLTabFormer~\cite{solatorio2023realtabformer} (RTF) are novel SOTA tabular data synthesizers based on pre-trained LLMs. 
By permuting the feature order during training, \great achieves to sample data conditioned on any given subset of features and samples the remaining features. 
Since both \great and \rtf adopt a fully text-based tokenization, they do not suffer from the dimension explosion stemming from encoding high-dimensional categorical columns. But their key disadvantage is the extremely long training time. 

\section{\alg Method}
In this section, we first explain the choices of foundation model for tabular data synthesis. Next we discuss on how to train and re-use pre-trained language models for new tasks. Then, we introduce a new token padding strategy specifically designed for tabular data synthesis.


\subsection{Foundation Model}
\label{ssec:foundation}

\begin{wraptable}{r}{0.6\textwidth}
\vspace{-2.3em}
\centering
\caption{Dataset Description. Abbreviations are in parentheses. Con. and Cat. represent the number of continuous and categorical columns.}
\resizebox{0.6\columnwidth}{!}{
\begin{tabular}{ |c|c|c|c|c| }
\hline
\textbf{Dataset} & \textbf{Task Type}& \textbf{Train/Test Split} & \textbf{$\mbox{Con.}$}  & \textbf{$\mbox{Cat.}$}\\
\hline
 Loan (LO)    & Binary & 4k/1k       &6&7\\\hline
{Adult (AD)}  & Binary  & 39k/9k   &5 &9\\
\hline
  Covertype (CO) & Multiclass& 40k/10k    &10 &45\\\hline
  Intrusion (IT) & Multiclass& 40k/10k    &22& 20\\\hline
  King (KI) & Regression &17k/4k&13&7\\
\hline
 Insurance (IS) & Regression &1k/300&3&4\\
 \hline
\end{tabular}
}
\vspace{-1.5em}
\label{table:DD}
\end{wraptable}

The choice between different LLMs depends on computational resources. While these LLMs are trained for natural language generation, tabular data synthesis requires transforming each data row into a sentence. Methods like \great use structures like "<column name> is <column value>," resembling natural language but forming distinct, repetitive patterns rarely seen in training datasets (i.e., BookCorpus~\cite{book} for GPT-2~\cite{Radford2019LanguageMA}). This domain mismatch reduces fine-tuning efficiency on pre-trained models. Instead, we propose using the LLMs with randomly initialized weights, enabling faster convergence by focusing solely on the unique patterns of tabular data synthesis.

\subsection{Re-usability of Pre-trained Model}
Recall that \great transforms each value "Y" in column "X" into a textual term "X is Y" for model training. In \alg, we simplify this term to "X Y", the reason is detailed in the next section.  
Training the model with this format familiarizes it with the pattern, enabling faster adaptation for new tasks.
Our tests, detailed in the experiment section, reveal that while most models pre-trained on tabular data outperform randomly initialized language models in new tabular data synthesis tasks, the extent of improvement varies. Their success lies in recognizing the "X Y" pattern, which encompasses both text order and data type. 
For robust performance, foundation models must be pre-trained on diverse data types (e.g., text, integers, decimals).
However, the scope for enhancement is not infinite. After mastering the pattern, to discern the relationships between X and Y, or among X, Y and other columns' values, the model requires further tuning for new tasks.

\subsection{Token Sequence Compression}
\begin{figure*}[t]
	\begin{center}
			\includegraphics[width=1\textwidth]{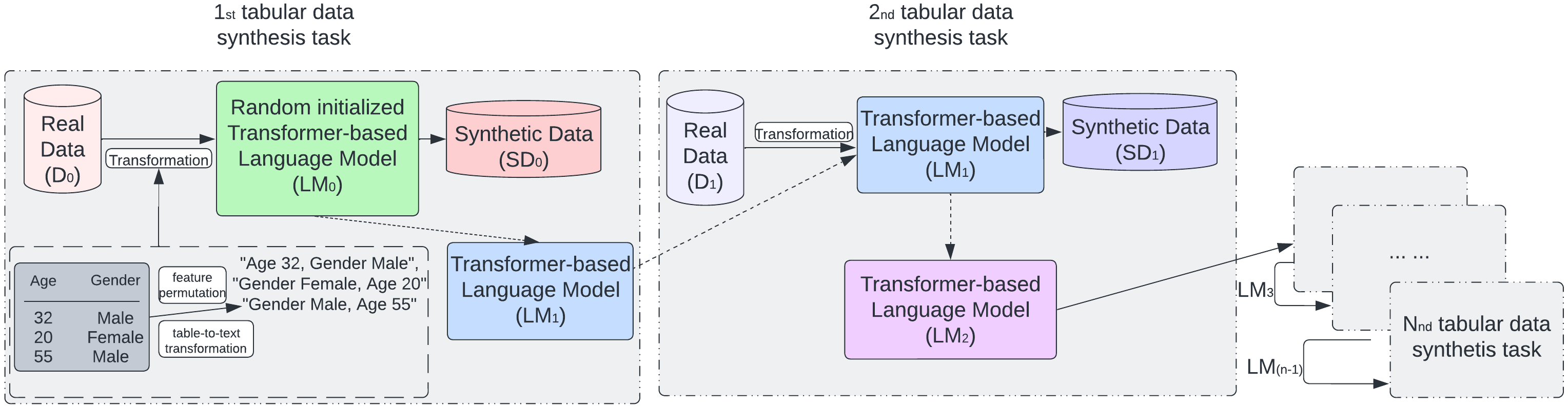}
		\caption{Initialization and Training Flow of \alg} 
		\label{fig:tabula}
 	\end{center}
  \vspace{-3em}
\end{figure*}

{
To optimize training speed, it is essential to minimize token sequence length. \alg employs the following pre-processing techniques:
(1) Token length reduction for column names and categorical values: evaluate the token lengths of all column names and values in categorical columns. Simplify these names and values to ensure they are tokenized into just one token. Column names and categorical values can be either abbreviated or substituted with a synonymous term. As illustrated in Fig.~\ref{fig:tabula}, when converting a table to a sentence, a single symbol suffices to represent the column name and the category value. This allows the LLM to correlate them with other values. Essentially, one indicator does the trick. It is important to ensure that any abbreviation or substitution is consistent across tables, enabling previously learned correlations by the model to be relevant for subsequent synthesis tasks.
(2) Simplified sentence transformation: while \great converts tables to text using the format "X is Y" (where 'X' denotes the column name and 'Y' represents its value), \alg streamlines this to just "X Y", omitting the word "is", as depicted in Fig.~\ref{fig:tabula}. \great's choice of "X is Y" stems from its foundation model, DistilGPT-2, which frequently encountered this structure in its training data, making it easier to learn. However, since \alg operates on a randomly initialized DistilGPT-2 model devoid of prior knowledge, the more concise "X Y" format is not only more efficient but also potentially simpler to learn due to its brevity.
By implementing these two pre-processing strategies, token sequence length can be sharply reduced compared to earlier methods.
}

\subsection{Middle Padding}



\great proposes feature order permutation during training, enabling conditional generation on any feature subset. However, this complicates model convergence due to the random feature order. To mitigate this, we introduce a novel token padding strategy when flexible conditional generation is unnecessary. \rtf maintains a fixed feature order during training and uses a fixed-set vocabulary tokenization method. While this approach prevents irrelevant token generation, its digit-by-digit numerical encoding disrupts value integrity and hinders the model's ability to capture inter-column value correlations.

\begin{figure}[t]
	\begin{center}
			\includegraphics[width=0.8\columnwidth]{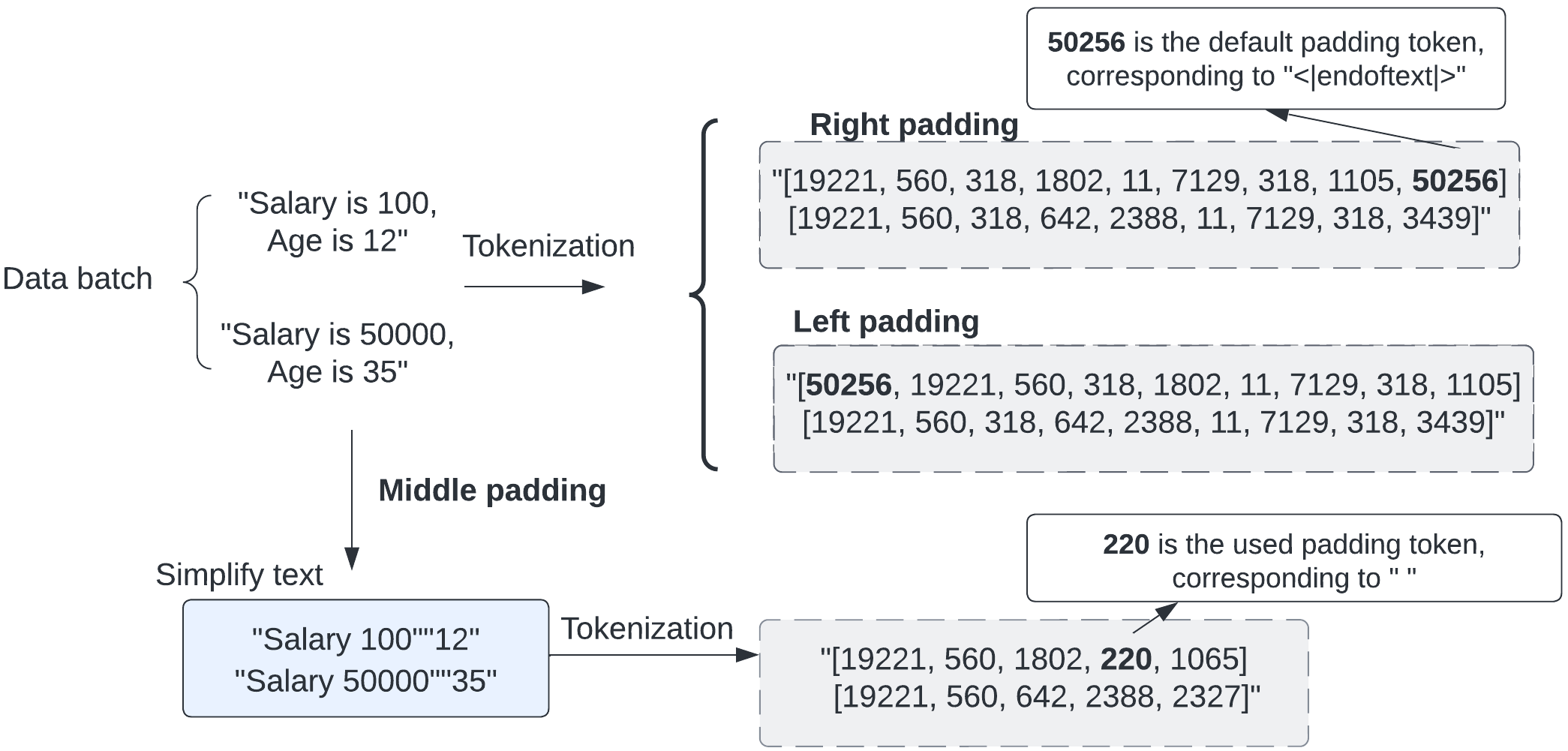}
		\caption{Middle Padding Strategy} 
		\label{fig:middle_padding}
 	\end{center}
  \vspace{-3em}
\end{figure}

The default GPT-2 tokenizer provides 2 padding modes: (1) right padding and (2) left padding. 
Fig.~\ref{fig:middle_padding} shows the process of left and right padding. When the sentences in the same batch tokenize to different sequence lengths, the tokenizer needs to add padding tokens (i.e., "50256" in Fig.~\ref{fig:middle_padding}) to the right or left of all shorter sequences to have all with equal length. For natural language, right or left padding is sufficient because tokens can relate to their preceding or succeeding tokens. 
However, by using our current table transformation method, distinct sentences now share a consistent token pattern. 
As a result, the absolute position of each token within the sequence holds structural significance. 
In the example shown in Fig.~\ref{fig:middle_padding}, right padding shifts the positions of the sub-token sequence "7129, 318" (i.e., tokens of "Age is"), while left padding misaligns the sequence "19221, 560, 318" (i.e., tokens of "Salary is"). This misalignment makes both padding strategies unsuitable for tabular data. Therefore, we propose the {middle padding} method.

Our approach extends beyond padding token sequences solely within a single data batch. Instead, we ensure alignment of token sequence lengths across the entirety of the dataset. 
We achieve this by firstly simplifying the text representation, as shown in Fig.~\ref{fig:middle_padding}.
For any given sentence, we retain only the primary column name. Subsequent data columns incorporate solely the data value, excluding the column name, and there are no spaces between column values. Following this, we segment the data representation column-wise and tokenize each column separately.
For each column, we find the longest length of the token sub-sequence specific to that column. Then during training, we consistently pad each sub-sequence to this pre-determined length. 
Retaining the initial column name serves a dual purpose: it acts as a starting prompt for data generation and mitigates issues arising from a potentially absent initial column value.
And the reason we only keep data values for the following columns is because since we consistently pad each column data into the uniform length and the data column order retains static for every data row, we can decode the generated tokens by their absolute positions for each column. We do not need their column name as the indicator anymore.
Such a method ensures that the token sub-sequence pertaining to each column  retains its absolute position throughout all token sequences, simultaneously reducing the token length. This refinement facilitates faster pattern recognition by the model, leading to reduced training time. 

\section{Experiment}
\subsection{Experimental Setup}
\label{ssec:setup}
{\bf Datasets}. All algorithms have been evaluated using six tabular datasets. The \textbf{Intrusion}, \textbf{Adult}, and \textbf{Covertype} datasets are sourced from the UCI ML Repository\footnote{\url{http://archive.ics.uci.edu/ml/datasets}}. The \textbf{Loan} dataset 
is obtained from Kaggle\footnote{\url{https://www.kaggle.com/bank-loan-modelling}}. These four tabular datasets feature a categorical variable as target, making them suitable for conducting classification tasks. To encompass regression tasks, two additional datasets, namely \textbf{Insurance} and \textbf{King}, sourced from Kaggle\footnote{\url{https://www.kaggle.com/{mirichoi0218/insurance,harlfoxem/housesalesprediction}}}, have been included. These datasets involve continuous target variables. Due to computational constraints, 50,000 rows are stratified randomly sampled from the Covertype and Intrusion datasets. The Adult, Loan, Insurance, and King datasets are used in full, with details in Table~\ref{table:DD}.

{\bf Baselines}. 
We evaluate \alg against five SOTA tabular synthesis algorithms (\ctgan, \ctabplus, \tabddpm, \great, and \rtf). While \ctgan, \ctabplus, and \tabddpm require prior data type knowledge, \great, \rtf, and \alg do not. All SOTA models and \alg are implemented in PyTorch using their original settings. GAN-based models are trained for 150 epochs (300 and 500 epochs for the Loan and Insurance datasets, respectively), while \great, \rtf, and \alg are trained for 50 epochs on larger datasets and 100 and 400 epochs for the Loan and Insurance datasets, respectively. Due to computational resource limitations, \great and \rtf adopt a pre-trained DistilGPT-2 model. \alg also uses the DistilGPT-2 model structure as its foundation framework; However, unless stated otherwise, its foundational model is a randomly initialized DistilGPT-2 fine-tuned on the Intrusion dataset. The Intrusion dataset is chosen because its categorical and continuous column counts are balanced, and it is relatively large among all the datasets; "Middle padding" is only employed in sec.~\ref{ssec:middle} when no feature permutation occurs during model training, in comparison with left padding, right padding. Above mentioned experiments are repeated 3 times and the average result with standard deviation is reported.

{\bf Environment}. Experimental machine equips with 32 GB memory, a GeForce RTX 3090 Ti GPU and a 10-core Intel i9 CPU under Ubuntu 20.04.

\begin{table*}[t]
\centering
\caption{ML utility result for synthetic data. For classification datasets, F1 score is reported. For regression dataset, MAPE is reported. Results are averaged over three runs with random seeds, best results are on \textbf{bold}.}
\resizebox{0.95\textwidth}{!}{%
\begin{tabular}{c c c c c c c c}
\hline
\textbf{Dataset} &  \textbf{Original} & \textbf{CTGAN}&  \textbf{CTABGAN+} &  \textbf{\great}  & \textbf{TabDDPM}  & \textbf{\alg}\\
\hline
Loan ($\uparrow$) & 0.929$\pm$.002&0.595$\pm$.006&0.812$\pm$.004&0.829$\pm$.003&0.751$\pm$.003&\textbf{0.902$\pm$.004}\\
\hline
Adult ($\uparrow$) & 0.723$\pm$.002&0.581$\pm$.004&0.687$\pm$.005&0.718$\pm$.003&0.719$\pm$.002&\textbf{0.740$\pm$.003}\\\hline
Covtype ($\uparrow$)& 0.777$\pm$.003&0.427$\pm$.007&0.636$\pm$.011&0.618$\pm$.003&\textbf{0.770$\pm$.002}&\textbf{0.770$\pm$.002}\\\hline
Intrusion ($\uparrow$)& 0.995$\pm$.001&0.805$\pm$.010&0.912$\pm$.004&0.977$\pm$.003&0.786$\pm$.005&\textbf{0.981$\pm$.002}\\\hline
King ($\downarrow$) & 0.255$\pm$.003&0.355$\pm$.009&0.277$\pm$.013&0.274$\pm$.006&0.282$\pm$.009&\textbf{0.250$\pm$.005}\\
\hline
Insurance ($\downarrow$)& 0.412$\pm$.006&0.516$\pm$.014& 0.467$\pm$.024&0.465$\pm$.009&0.517$\pm$.007&\textbf{0.430$\pm$.008}\\\hline
\end{tabular}
} 
\label{table:ml_utility}
\vspace{-1em}
\end{table*}

\subsection{Evaluation Metrics}
The synthetic data evaluation encompasses two key facets: (1) \textbf{machine learning utility}: We evaluate classification and regression datasets using a common process. Original data is split into 80\% training and 20\% test sets. Models are separately trained on both real and synthetic data, and their performance is evaluated on the test set. For classification, we use decision trees, support-vector-machine (SVM), random forest, logistic regression, and multilayer perceptron (MLP), measuring F1-score. For regression, we use linear, ridge, lasso, and Bayesian ridge regression, evaluating with mean absolute percentage error (MAPE). Average scores across models are reported per dataset. (2) \textbf{Statistical similarity}:
We assess the faithfulness of column dependencies in synthetic data by calculating pair-wise correlation matrices for real and synthetic datasets. Continuous variables are evaluated using the \textit{Pearson correlation} coefficient ($[-1, +1]$), categorical features with the \textit{uncertainty coefficient} ($[0, 1]$), and categorical-continuous relationships with the \textit{correlation ratio} ($[0, 1]$). Using the dython library\footnote{\url{http://shakedzy.xyz/dython/modules/nominal/\#compute_associations}}, we compute the Correlation Distance, where lower values indicate higher synthesis quality.
Notably, \textbf{a lower correlation distance value indicates higher synthesis quality}.

\subsection{Result Analysis}
\subsubsection{Foundation model choice and re-usability}
\label{ssec:continuous}


Tab.~\ref{table:ml_utility} shows the ML utility results for baseline methods and \alg. \rtf, which lacks feature permutation during training, is excluded and compared separately to \alg with middle padding in Tab.~\ref{table:middle_padding}. Compared to \great, \alg differs in foundation models and data representation, consistently outperforming \great and all other baselines across datasets. Notably, on the Adult and King datasets, \alg's synthetic data achieves higher ML utility than the original data, highlighting its ability to emulate and understand the original data's distribution.

\subsubsection{Ablation study} 
\begin{figure}[t]\centering
   \begin{minipage}{0.37\columnwidth}
    \vspace{0em}
     \includegraphics[width=1\linewidth]{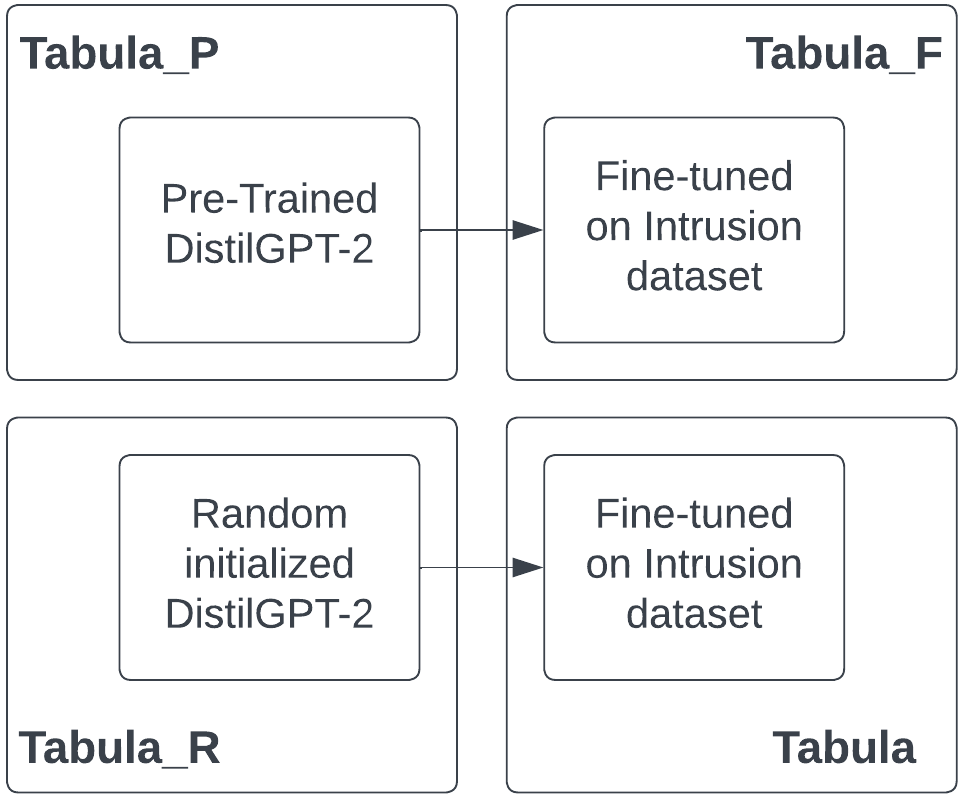}
     \caption{Relations among \algp, \algf, \algr and \alg.}
     \label{fig:fd_model}
   \end{minipage}
   \hspace{0.4em}
   \begin {minipage}{0.43\columnwidth}
     \includegraphics[width=1\linewidth]{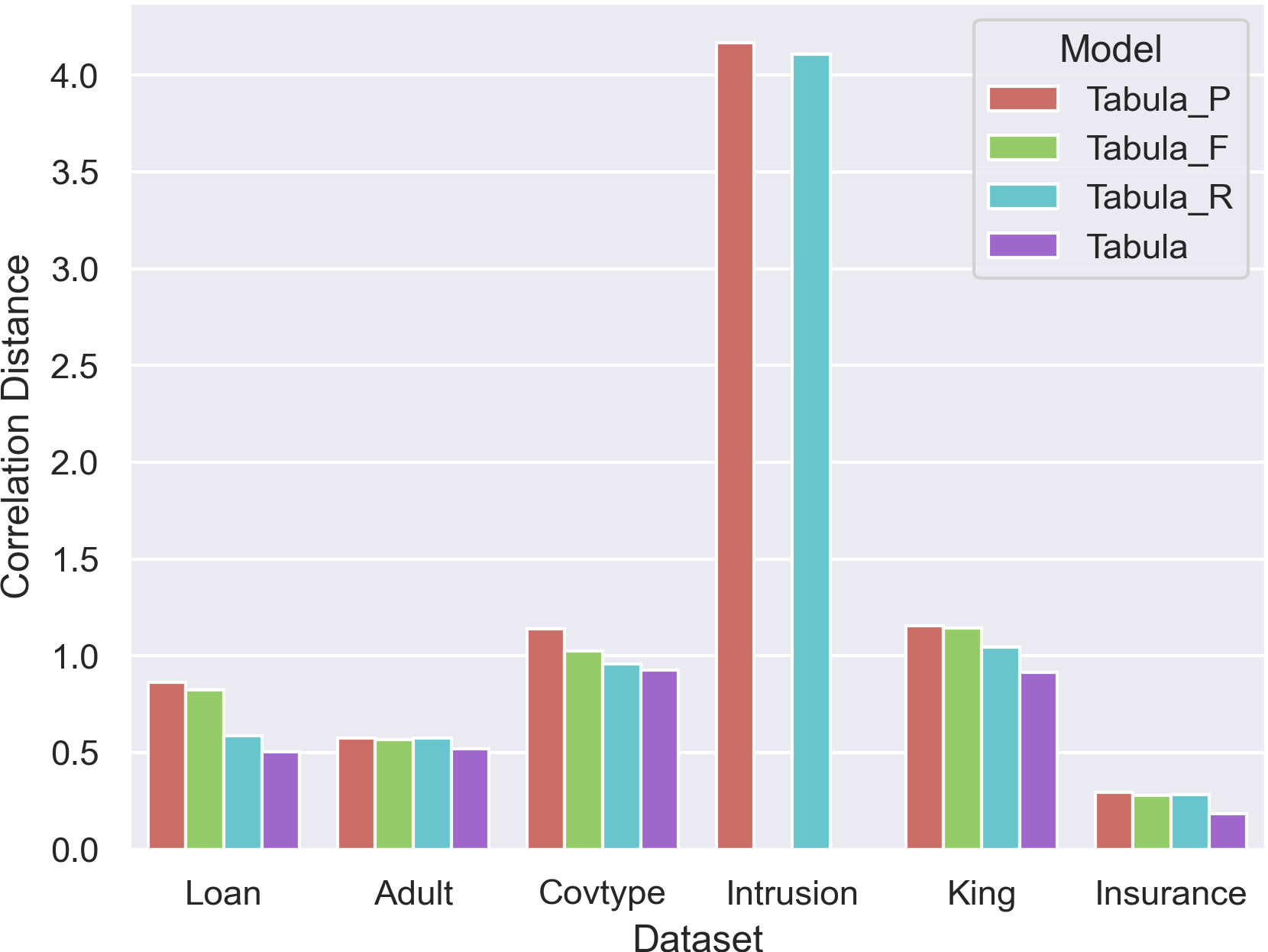}
     \vspace{-1.5em}
     \caption{Correlation distance result for \algp, \algf, \algr and \alg.}
     \label{fig:correlation_distance}
   \end{minipage}
   \vspace{-2em}
\end{figure}

{We define a model \textbf{\algp} which uses pre-trained DistilGPT-2 as the foundation model for \alg. We then use the \textbf{\algp} for fine-tuning on Intrusion dataset synthesis task, yielding a saved fine-tuned model.  Here, \textbf{\algf} designates the variant of the \alg algorithm where the foundation model is replaced with the Intrusion fine-tuned model trained from pre-trained DistilGPT-2. 
Recall that \alg in this paper uses the foundation model which is re-used from training for Intrusion dataset synthesis. Then we introduce \textbf{\algr} as the configuration wherein \alg employs a randomly initialized DistilGPT-2 model as its foundation. 
Foundation model relations for \algp, \algf, \algr and \alg are shown in Fig.~\ref{fig:fd_model}. }
Fig.~\ref{fig:correlation_distance} shows the correlation distance achieved across all six datasets for \algp, \algf, \algr, and \alg. A consistent pattern emerges: \algf achieves outperforms \algp, and \alg outperforms \algr. This highlights the importance of fine-tuning on the Intrusion dataset for better performance, regardless of whether the starting model is pre-trained or randomly initialized. Additionally, \algr consistently surpasses \algp, demonstrating that a randomly initialized DistilGPT-2 outperforms the default pre-trained version for tabular data synthesis. Finally, \alg achieves the best performance on all datasets, underscoring the synergy between random initialization and fine-tuning, which enhances the foundation model for tabular data synthesis.



\begin{wraptable}{r}{0.7\textwidth}
\vspace{-2.3em}
   \begin{minipage}{0.35\columnwidth}
     \includegraphics[width=1\linewidth]{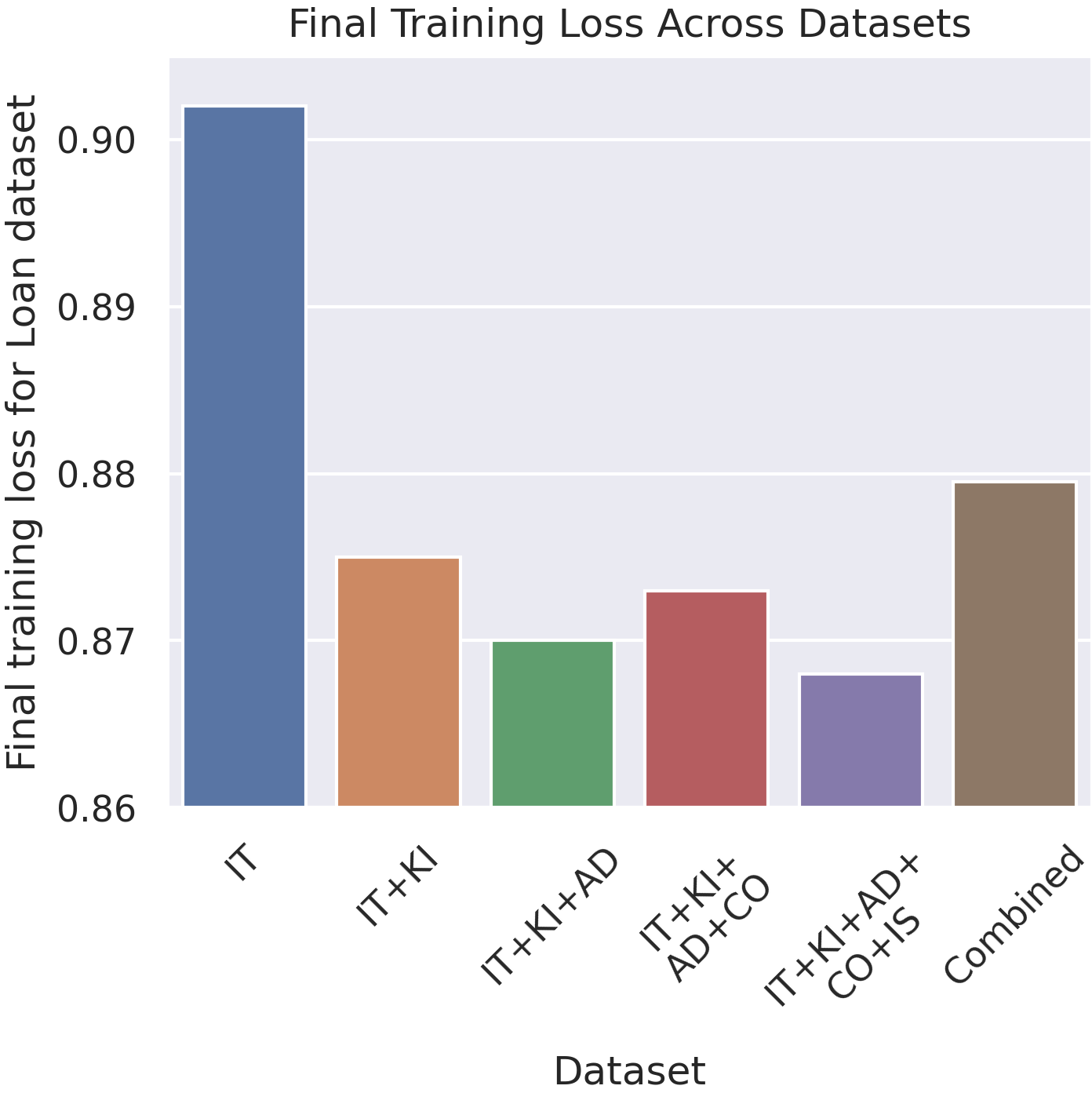}
     \vspace{-1.4em}
     \caption{\small{Final training loss on Loan dataset with foundation model iteratively trained on different datasets.}}
     \label{fig:accumulated_loss}
   \end{minipage}
   \begin {minipage}{0.3\columnwidth}
     \includegraphics[width=1\linewidth]{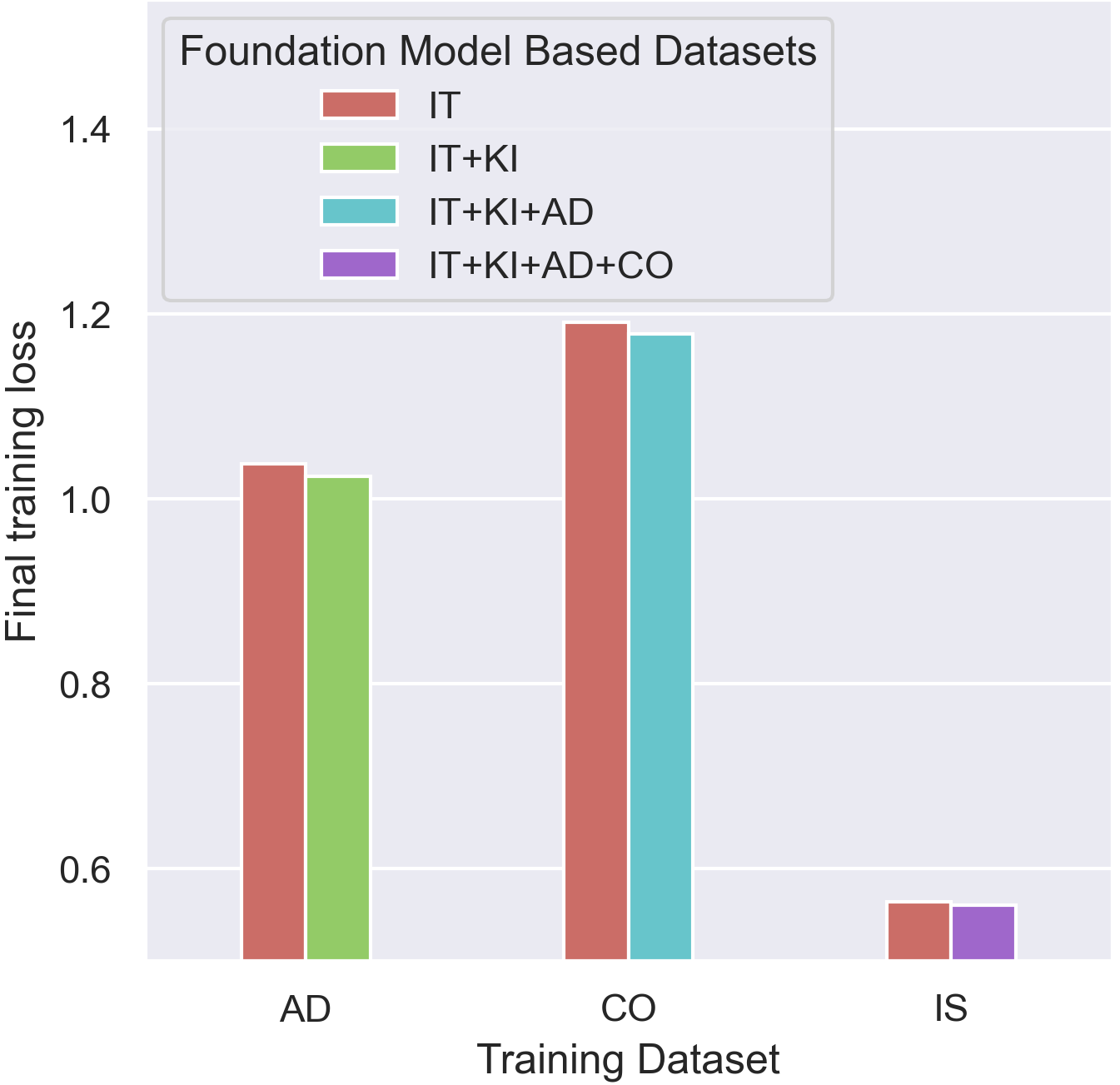}
     \vspace{0.3em}
     \caption{\small{Final training loss with different foundation model. Dataset abbreviation is in Tab.~\ref{table:DD}.}}
     \label{fig:accumulated_loss_continuous}
   \end{minipage}
   \vspace{-1.8em}
\end{wraptable}

\subsection{Further improvements to foundation model}




Previous experiments highlight the ability of a fine-tuned model, originating from the Intrusion dataset, to accelerate convergence on new tabular data synthesis tasks. Building on this, we design an experiment involving iterative fine-tuning across new datasets.
The order of these tasks is determined through an experiment where models were first fine-tuned separately on all datasets except Loan. Each fine-tuned model was then further fine-tuned on the Loan dataset. 
Ranking the synthetic Loan dataset quality from high to low from these experiments gives the order: Intrusion, King, Adult, Covtype, and Insurance (Details omitted due to space constraints.). Following this order, we fine-tune the model iteratively, saving intermediate models after each task.
Meanwhile, we conduct an experiment combining data from Intrusion, King, Adult, Covtype, and Insurance into a single dataset, transforming each tabular row into a sentence. This consolidated dataset is used to fine-tune the model in one cycle. 
Subsequently, all the above models are employed as the foundation model for further fine-tuning on the Loan dataset synthesis task.  
Fig.~\ref{fig:accumulated_loss} illustrates the final training loss of fine-tuning tasks using each intermediate model, with results displayed in fine-tuning order on the x-axis. \textit{Combined} represents the outcome of fine-tuning on all datasets simultaneously.
Three key observations emerge:  
(1) Iterative fine-tuning demonstrates the foundation model's capacity for continuous improvement, with the last four models achieving lower final losses than the initial one, though the trend is not strictly monotonic.
(2) The model fine-tuned with Intrusion, King, and Adult datasets (3rd column) outperforms the one fine-tuned with Intrusion, King, Adult, and Covtype datasets (4th column) on the Loan dataset synthesis task.
(3) While \textit{Combined} does not achieve the best results, it still performs well, offering more generalizable outcomes as it operates independently of the fine-tuning order.
These findings highlight the value of pre-training on tabular data for building robust foundation models.
To demonstrate the extended effectiveness of the refined foundation model beyond the Loan dataset, we evaluate its impact on other datasets. Following the fine-tuning sequence in Fig.~\ref{fig:accumulated_loss}, we track the final training loss at each step and compare it to the default \alg configuration (i.e., the Intrusion dataset fine-tuned model).
Fig.~\ref{fig:accumulated_loss_continuous} presents these results. The "IT" bar represents the default \alg setting, while the bar on the right shows the final loss of the new foundation model, with the legend indicating the fine-tuned datasets. The figure demonstrates that successive fine-tuning iterations significantly accelerate the convergence of synthesis tasks.

\color{black}

\subsubsection{Token Sequence Compression}
\begin{wraptable}{r}{0.6\textwidth}
\vspace{-2.3em}
\caption{The performance changes without token sequence compression.}
\centering
\resizebox{0.6\columnwidth}{!}{%
\begin{tabular}{l c c c c c c}
\hline
Metrics     & Loan & Adult & Covtype & Intrusion & King & Insurance \\ \hline
F1-score ($\uparrow$) & -1.3\%    &   -1.1\%    & -4.1\%         &      -1.1\%     & -1.0\%      &  -3.9\%          \\ \hline
Corr. Diff. ($\downarrow$) &  +15.3\%    &   +4.4\%    &   +6.2\%      &      +0.2\%     &    +10.6\%  &  +18.1\%          \\ \hline
\end{tabular}
}
\vspace{-2em}
\label{tab:ablation_rename}
\end{wraptable}
Incorporating the token sequence compression strategy yields an intuitive advantage of reducing the training time per epoch. To comprehensively assess the impact of this strategy on synthetic data quality, we conduct an ablation test on the \alg algorithm. This evaluation temporarily disables the token sequence compression within \alg, retaining the table-to-text transformation proposed by \great. The performance differences are detailed in Tab.~\ref{tab:ablation_rename}. 
The result reveals a consistent trend: disabling the token sequence compression method reduces ML utility (lower F1-Score) and worsens statistical similarity (higher Corr. Diff.) between synthetic and real data. This highlights how token sequence compression shortens token sequences, simplifies learning, and improves training performance.
\color{black}

\subsubsection{Middle padding strategy}
\label{ssec:middle}
\begin{wraptable}{r}{0.65\textwidth}
\vspace{0em}
\centering
\caption{ML utility result without feature permutation during training.}
\resizebox{0.65\columnwidth}{!}{%
\begin{tabular}{c c c c c}
\hline
\textbf{Dataset} &\textbf{\rtf} & \textbf{\algll} &  \textbf{\algrr} & \textbf{\algmm}\\
\hline
Loan ($\uparrow$)&0.900$\pm$.001& 0.880$\pm$.003 & 0.884$\pm$.001 & \textbf{0.920$\pm$.001}\\
    \hline
Adult ($\uparrow$) &0.704$\pm$.002& 0.738$\pm$.004& 0.729$\pm$.003&\textbf{0.755$\pm$.003}\\
\hline
Covtype ($\uparrow$) &0.760$\pm$.002&0.765$\pm$.002&0.769$\pm$.002&\textbf{0.770$\pm$.002}\\
\hline
Intrusion ($\uparrow$)& {0.981$\pm$.001}&0.963$\pm$.002&0.971$\pm$.001&\textbf{0.984$\pm$.001}\\
\hline
King ($\downarrow$) &0.264$\pm$.004&0.280$\pm$.004&0.311$\pm$.004&\textbf{0.245$\pm$.003}\\
\hline
Insurance ($\downarrow$)&\textbf{0.412$\pm$.004}&0.502$\pm$.006&0.422$\pm$.006&\textbf{0.412$\pm$.005} \\
\hline
\end{tabular}
}
\label{table:middle_padding}
\vspace{-2em}
\end{wraptable} 
To demonstrate the effectiveness of middle padding in tabular data synthesis, we compare three \alg variants: \algll (left padding), \algrr (right padding), and \algmm (middle padding), all without feature permutation during training. We also benchmark them against \rtf, which uses a fixed feature order. Results in Tab.\ref{table:middle_padding} show that \algmm outperforms the other padding strategies and \rtf on five out of six datasets, matching \rtf on the sixth. Notably, \algmm even surpasses \alg in Tab.\ref{table:ml_utility} for five out of six datasets, particularly those with fewer features. With fewer features, there are fewer correlations to learn. While DistilGPT-2’s autoregressive nature introduces some artificial dependencies by the fixed feature order, the model better captures broader column correlations, enhancing performance.

\subsubsection{Training Time Analysis}
\color{black}
After evaluating synthetic data quality, we analyze training times for various baseline models. Tab.~\ref{table:training_time} shows training time per epoch across datasets. GAN-based synthesizers train significantly faster than LLM- or diffusion-based algorithms. Among LLM-based methods, \alg achieves a 46.2\% average reduction in training time compared to \great. Notably, both without feature permutation, \rtf trains slightly slower than \algmm due to its digit-by-digit encoding of numerical values.
\begin{table}[t]
\centering
\caption{Training Time (s/epoch) Usage.}
\resizebox{0.9\columnwidth}{!}{%
\begin{tabular}{c c c c c c | c c}
\hline
\textbf{Dataset} & \textbf{CTGAN}&  \textbf{CTABGAN+} &  \textbf{\great}  & \textbf{TabDDPM}  & \textbf{\alg} & \textbf{\rtf} & \textbf{\algmm}\\
\hline
Loan  &0.2 &0.4&13.9&101.5&7.9&4.9&4.5\\
\hline
Adult &1.8 &11.1&156.1&153.6&104.2&93.3&86.1\\\hline
Covtype &3.4 &9.2&854.1&898.1&379.2&219.0&216.1\\\hline
Intrusion & 3.1&12.5&835.5&839.3&406.2&286.5&238.0\\
\hline
King &1.0&5.3&123.1&121.2&75.8&76.5&59.1\\\hline
Insurance &0.1&0.4&4.7&103.2&2.1&1.4&1.0\\\hline
\end{tabular}
} 
\vspace{-1em}
\label{table:training_time}
\end{table}

\begin{wraptable}{r}{0.42\textwidth}
\vspace{-2.3em}
\centering
\caption{Maximal Token Sequence Length of One Data Row.}
\resizebox{0.4\columnwidth}{!}{%
\begin{tabular}{c c c | c c}
\hline
\textbf{Dataset} &\textbf{\great} & \textbf{\alg} &  \textbf{\rtf} & \textbf{\algmm}\\
\hline
Loan &62	&41&	27&	\textbf{18}\\
    \hline
Adult &74&	44	&34	&\textbf{21}\\
\hline
Covtype  &447&	177	&81&	\textbf{63}\\
\hline
Intrusion & 378	&162&	121&	\textbf{80}\\
\hline
King  &126	&78&	84&	\textbf{46}\\
\hline
Insurance &36	&27	&24	&\textbf{16} \\
\hline
\end{tabular}
}
\vspace{-2em}
\label{table:token_sequence_length}
\end{wraptable}

To explain variations in training times for LLM-based methods, Tab.\ref{table:token_sequence_length} shows token sequence lengths for a representative data row across algorithms. \alg achieves significant compression compared to \great, with reductions of up to 60\% for Covtype and 57\% for Intrusion. Even though \rtf already reduces token sequence lengths, \algmm compresses them further. However, similar token sequence lengths between the Loan and Adult datasets do not result in comparable training times per epoch in Tab.\ref{table:training_time}. This discrepancy stems from dataset size: the Adult dataset has approximately ten times more samples than the Loan dataset (Tab.~\ref{table:DD}), driving the difference in training times.

\color{black}

\section{Acknowledgement}
This work was supported in part by the Asian Institute of Digital Finance (AIDF) under grant A-0003504-09-00.
\section{Conclusion}

{

This paper presents \alg, a novel tabular data synthesis algorithm leveraging large language models (LLMs) to address the challenge of long training times.
We challenge the reliance on natural language processing (NLP)-optimized pre-trained models, advocating for randomly initialized models as more effective starting points. Through iterative fine-tuning, we show that continuously refined models become efficient foundations for successive tasks. Additionally, our token sequence compression simplifies data representation, improving performance and reducing complexity. Finally, the proposed middle padding strategy outperforms existing methods, enhancing training efficiency for fixed feature order scenarios.
\bibliographystyle{abbrv}
\bibliography{main}

\begin{thebibliography}{10}

\bibitem{avino2018generating}
L.~Avi{\~n}{\'o}, M.~Ruffini, and R.~Gavald{\`a}.
\newblock Generating synthetic but plausible healthcare record datasets.
\newblock {\em arXiv preprint arXiv:1807.01514}, 2018.

\bibitem{borisov2022language}
V.~Borisov, K.~Sessler, T.~Leemann, M.~Pawelczyk, and G.~Kasneci.
\newblock Language models are realistic tabular data generators.
\newblock In {\em The Eleventh International Conference on Learning Representations}, 2023.

\bibitem{sos}
J.~Kim, C.~Lee, Y.~Shin, S.~Park, M.~Kim, N.~Park, and J.~Cho.
\newblock Sos: Score-based oversampling for tabular data.
\newblock In {\em Proceedings of the 28th ACM SIGKDD Conference on Knowledge Discovery and Data Mining}, KDD '22, page 762–772, New York, NY, USA, 2022. Association for Computing Machinery.

\bibitem{kotelnikov2023tabddpm}
A.~Kotelnikov, D.~Baranchuk, I.~Rubachev, and A.~Babenko.
\newblock Tabddpm: Modelling tabular data with diffusion models.
\newblock In {\em International Conference on Machine Learning}, pages 17564--17579. PMLR, 2023.

\bibitem{itgan}
J.~Lee, J.~Hyeong, J.~Jeon, N.~Park, and J.~Cho.
\newblock Invertible tabular gans: Killing two birds with one stone for tabular data synthesis.
\newblock {\em Advances in Neural Information Processing Systems}, 34:4263--4273, 2021.

\bibitem{nowok2016synthpop}
B.~Nowok, G.~M. Raab, and C.~Dibben.
\newblock synthpop: Bespoke creation of synthetic data in r.
\newblock {\em Journal of statistical software}, 74:1--26, 2016.

\bibitem{copulas}
T.~S. D.~V. Project.
\newblock Copulas, 2022.

\bibitem{Radford2019LanguageMA}
A.~Radford, J.~Wu, R.~Child, D.~Luan, D.~Amodei, and I.~Sutskever.
\newblock Language models are unsupervised multitask learners.
\newblock 2019.

\bibitem{solatorio2023realtabformer}
A.~V. Solatorio and O.~Dupriez.
\newblock Realtabformer: Generating realistic relational and tabular data using transformers.
\newblock {\em arXiv preprint arXiv:2302.02041}, 2023.

\bibitem{ctgan}
L.~Xu, M.~Skoularidou, A.~Cuesta-Infante, and K.~Veeramachaneni.
\newblock Modeling tabular data using conditional gan.
\newblock In {\em Advances in Neural Information Processing Systems, 2019}, volume~32, pages 7335--7345. Curran Associates, Inc., 2019.

\bibitem{privb}
J.~Zhang, G.~Cormode, C.~M. Procopiuc, D.~Srivastava, and X.~Xiao.
\newblock Privbayes: Private data release via bayesian networks.
\newblock {\em ACM Trans. Database Syst.}, 42(4), oct 2017.

\bibitem{ctabgan}
Z.~Zhao, A.~Kunar, R.~Birke, and L.~Y. Chen.
\newblock Ctab-gan: Effective table data synthesizing.
\newblock In {\em Proceedings of The 13th Asian Conference on Machine Learning}, volume 157, pages 97--112, 17--19 Nov 2021.

\bibitem{ctabplus}
Z.~Zhao, A.~Kunar, R.~Birke, and L.~Y. Chen.
\newblock Ctab-gan+: Enhancing tabular data synthesis.
\newblock {\em arXiv preprint arXiv:2204.00401}, 2022.

\bibitem{book}
Y.~Zhu, R.~Kiros, R.~Zemel, R.~Salakhutdinov, R.~Urtasun, A.~Torralba, and S.~Fidler.
\newblock Aligning books and movies: Towards story-like visual explanations by watching movies and reading books.
\newblock In {\em The IEEE International Conference on Computer Vision (ICCV)}, December 2015.

\bibitem{zhu2022}
Y.~Zhu, Z.~Zhao, R.~Birke, and L.~Y. Chen.
\newblock Permutation-invariant tabular data synthesis.
\newblock In {\em 2022 IEEE International Conference on Big Data (Big Data)}, pages 5855--5864, 2022.

\end{thebibliography}

\end{document}